# CTL-MTNet: A Novel CapsNet and Transfer Learning-Based Mixed Task Net for the Single-Corpus and Cross-Corpus Speech Emotion Recognition

Xin-Cheng Wen[1*], Jia-Xin Ye[1*], Yan Luo[1], Yong Xu[2], Xuan-Ze Wang[1],
Chang-Li Wu[1] and Kun-Hong Liu[3,1†]

[1]School of Informatics, Xiamen University, Xiamen, China
[2]School of Computer Science and Mathematics, Fujian University of Technology, Fuzhou, China
[3]School of Film, Xiamen University, Xiamen, China
{xiamenwxc, jiaxin-ye, lauren-ly, xuanze.wang, changli-wu} @foxmail.com,
y.xu@fjut.edu.cn, lkhqz@xmu.edu.cn

## Abstract

Speech Emotion Recognition (SER) has become a growing focus of research in human-computer interaction. An essential challenge in SER is to extract common attributes from different speakers or languages, especially when a specific source corpus has to be trained to recognize the unknown data coming from another speech corpus. To address this challenge, a Capsule Network (CapsNet) and Transfer Learning based Mixed Task Net (CTL-MTNet) are proposed to deal with both the single-corpus and cross-corpus SER tasks simultaneously in this paper. For the single-corpus task, the combination of Convolution-Pooling and Attention CapsNet module (CPAC) is designed by embedding the self-attention mechanism to the CapsNet, guiding the module to focus on the important features that can be fed into different capsules. The extracted high-level features by CPAC provide sufficient discriminative ability. Furthermore, to handle the cross-corpus task, CTL-MTNet employs a Corpus Adaptation Adversarial Module (CAAM) by combining CPAC with Margin Disparity Discrepancy (MDD), which can learn the domain-invariant emotion representations through extracting the strong emotion commonness. Experiments including ablation studies and visualizations on both single- and cross-corpus tasks using four well-known SER datasets in different languages are conducted for performance evaluation and comparison. The results indicate that in both tasks the CTL-MTNet showed better performance in all cases compared to a number of state-of-the-art methods. The source code and the supplementary materials are available at: https://github.com/MLDMXM2017/CTL-MTNet.

## 1 Introduction

As an important communication way among people, speech conveys emotion, pitch and rhythm to express their thoughts and intentions to others. The understanding of emotional information in a speech is important in predicting the speaker's tendencies and reactions. The prediction of human emotions from speech signals is known as the Speech Emotion Recognition (SER) task, which has been attracting more and more researchers' attention in the human-computer interaction research field.

In the SER task, different speech signals collected from various speakers tend to exhibit diverse characteristics, but there are implicit emotional attributes in the same emotion category. Therefore, the essential challenge in SER is to discover such common representations so as to build robust models to predict human speech emotions. Most of such models were developed based on Convolutional Neural Networks (CNN) and Long Short-Term Memory (LSTM) networks [Akçay and Oğuz, 2020] with manually designed features to extract local spatial and temporal information from speech. Verbitskiy *et al.*[2021] designed a CNN architecture by carefully adjusting its hyperparameters to reduce the size of the SER model significantly. In addition, the Capsule Neural Network (CapsNet) [Sabour *et al.*, 2017] was also first used by Wu *et al.*[2019] in the SER domain to process the spatial relationships of speech features in spectrograms due to its effectiveness in extracting spatial information. Liu *et al.*[2020] further embedded CapsNet into a local-global aware deep representation learning system to perform emotion classification. All these methods could lead to high performance on the SER task.

However, these SER methods were proposed without considering the scenarios where the training data (source corpus) and test data (target corpus) come from different speech corpora. In this case, the huge gap in the model's performance between the source and target corpora was ignored. This gap is mainly caused by the speech signals collected from different language environments, such as various vocal environments or different ways of expressing emotions. Consequently, different from the single-corpus task, which only needs to split the training and test data from the same corpus, the cross-corpus SER task tends to distinguish the training

---

and test speech signals coming from different corpora. The weak potential correlation of data distributions between the source and target corpora makes the cross-corpus task more challenging than the single-corpus task.

In the cross-corpus task, researchers tried to unlabel the agglomerate data and align features from diversified spaces [Zhang *et al.*, 2021]. For example, Domain Adversarial Neural Network (DANN) was proposed to efficiently align mismatched feature distributions [Abdelwahab and Busso, 2018]. A non-negative matrix factorization based transfer subspace learning method [Luo and Han, 2020] was designed to search for an optimally shared feature subspace for the source and target corpora, aiming to better eliminate the discrepancy between two distributions. Furthermore, the Variational Auto-encoding Wasserstein Generative Adversarial Network (VAW-GAN) implemented an emotional style transfer framework to bridge the signals across different data sets [Zhou *et al.*, 2021].

Nevertheless, there are still some limitations to the proposed methods for the SER task, including:

1. There are no known algorithms that suit both the single-corpus and cross-corpus SER tasks simultaneously.

2. Most methods utilized complex hand-crafted Low-Level Descriptors as inputs [Schuller *et al.*, 2010], in which the extraction of salient features was highly dependent on the training data.

3. The mainstream transfer learning methods in SER can only learn 0-1 loss, which is not suitable to handle the multi-class classification problems related to the cross-corpus task directly.

To address these challenges, this paper proposes a CapsNet incorporated with the Transfer Learning-based Mixed Task Net (CTL-MTNet) to tackle the single- and cross-corpus SER tasks simultaneously. It is noted that in the traditional CapsNet, each capsule individually represents some properties of the entity, which require proper strategies to dynamically evaluate their importance. Therefore, for the single-corpus task, the module is designed through the combination of Convolution-Pooling and Attention-based CapsNet (CPAC for short) to enhance the effectiveness of important capsules. CPAC calculates the emotion information scores to adjust the attention weights for each capsule entity accordingly. In this way, larger weights are assigned to the more informative capsule entities to promote the discriminative ability.

Besides, the Corpus Adaptation Adversarial Module (CAAM) is adopted in the cross-corpus task with the most commonly used Mel-Frequency Cepstral Coefficients (MFCCs) features. CAAM employs the Margin Disparity Discrepancy (MDD) loss [Zhang *et al.*, 2019] and the cross-entropy loss to balance the prediction results on sentiment feature alignment and emotion discrimination. The combination of CPAC and CAAM guarantees the high performance of our framework in both single- and cross-corpus SER tasks.

The main contributions are summarized as follows:

**The first framework for handling two SER tasks simultaneously.** To the best of our knowledge, CTL-MTNet is the first model that can handle the single-corpus and cross-corpus tasks simultaneously. It can not only extract the common features of the source corpus and the target corpus for the cross-corpus task, but also extract the individual emotional characteristics in the speech for the single-corpus task.

**A novel high-level feature extraction component.** The proposed CPAC mechanism inherits CapsNet's ability to capture local relevance and global contextual information, and generates high-level features. In addition, it can also identify important features contained in various capsules through the attention mechanism.

**The design of CAAM module for the cross-corpus task.** The CAAM module is embedded in our framework with the aid of MDD to learn the domain-invariant emotion representations with high robustness to bridge the gap between different corpora.

## 2 Methodology
### 2.1 The Framework of CTL-MTNet

As is shown in Fig.1, our model employs CPAC for the single-corpus task, and CAAM for the cross-corpus task. MFCC features are extracted from audio data to serve as input features. The CNN-Pooling module generates relevant emotion features by capturing local correlations from MFCC features. The Attention CapsNet module consists of a PrimaryCaps layer, a self-attention layer and a DigitCaps layer. The PrimaryCaps layer adjusts each sample from a scalar neuron to a vector neuron layer. The vector length is used to represent the probability of the capsule entity.

Furthermore, the self-attention layer calculates the correlation scores among capsules, aiming to guide the module to focus on the connections across different capsules at the DigitCaps layer. The attentional representations learned by the DigitCaps are iterated through the capsule routing algorithm to produce robust and discriminative high-level speech features, which are sequentially fed to the global average pooling layer. The final prediction is generated based on the softmax function in the single-corpus task.

Besides, the proposed algorithm employs the CAAM module to effectively deal with the cross-corpus task by extracting common speech features. For the MFCC features of the source corpus and the target corpus, CPAC extracts high-level features with discriminative emotion-preserving representations, and MDD is used to further learn the domain-invariant emotion representation via feature alignment. After successive iterations, CAAM gradually generates representations with more discriminative emotion-preserving and a minor discrepancy between corpora.

### 2.2 CNN-Pooling Feature Learning

As is shown in Fig.1, three one-dimensional convolution blocks are connected to extract high-level features from MFCC further. Each CNN-pooling block consists of a 2D convolution operation followed by batch normalization, *elu* activation function, and an average pooling layer with a fixed dropout at a rate of 0.25. The kernel size of all convolutional layers is $[3 \times 3]$, and the number of filters is 64. The CNN-pooling blocks act as a mapping of the MFCC features to a high-dimensional space for training capsules.

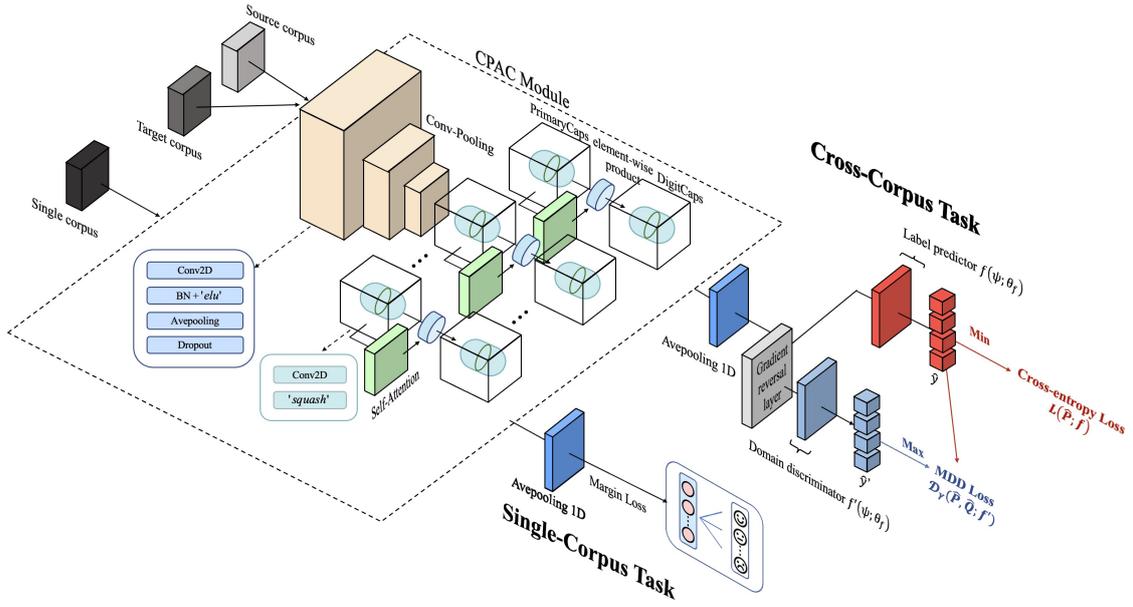

Figure 1: The architecture of CTL-MTNet. The upper and lower parts represent the cross-corpus task and the single-corpus task, respectively.

### 2.3 Attention CapsNet

Attentional vocalization in speech has a more significant role in perceiving human emotions. As is shown in Sabour *et al.*[2017], the state and properties of each capsule include many different types of instantiated parameters. Our method attempts to establish a relationship among different capsules by the attention mechanism. The output of the CNN-Pooling module is fed to the PrimaryCaps layer, which consists of a 2D convolution operation and the Squash function. The output of the PrimaryCaps layer is the product of the capsule dimension and the number of target capsules. It is equivalent to stacking multiple parallel conventional layers. The length of the input and output vectors of CapsNet represents the probability of an emotional entity. The Squash nonlinear function is employed as the activation function:

$$v_j = \frac{||s_j||^2}{1 + ||s_j||^2} \cdot \frac{s_j}{||s_j||} \quad (1)$$

Where $j$ represents the specified capsule and $s_j$ represents the output of the convolution layer within the PrimaryCaps. The squash function ensures that the length of the short vector is reduced to almost zero, while the length of the long vector is scaled to close to but no more than one. The output of the PrimaryCaps layer is also adjusted to fit the vector neuron layer and is mapped to three identical linear mappings $A^Q$, $A^K$, $A^V$. These mappings represent the queries, keys, and values of the self-attention mechanism, which all have the same dimension $d$. The output attention is then calculated by:

$$Attention(A^Q, A^K, A^V) = softmax\left(\frac{A^Q {A^K}^T}{\sqrt{d}}\right) A^V \quad (2)$$

With the aid of the attention mechanism, our CapsNet can further extract spatial information from the inputs, and establish connections for the representations extracted from the capsules. It provides distinguishing feature learning results for subsequent components. Moreover, the network computes the similarity among the sequence information to mine the emotional representation in the capsules. The input $I_{dig}$ of the DigitCaps layer is obtained from the attention scores and the output $O_{pri}$ of the PrimaryCaps layer:

$$I_{dig} = Attention(A^Q, A^K, A^V) \times O_{pri} \quad (3)$$

Then, the output is fed to the DigitCaps and Avepooling layers to further extract high-level speech features. The self-attention score is gradually improved through iterative refinement to establish the connections among different capsules. This approach provides more robust and discriminative representations for the subsequent operations in both single-corpus and cross-corpus tasks, thus improving the overall performance of our algorithm.

### 2.4 Margin Loss

In the single-corpus task, the output layer uses a vector other than scalar representation. The length of each vector represents the probability that the capsule entity exists. Since there are multiple emotions classes, the overall margin loss is the sum of all samples' loss, calculated by:

$$L_k = T_k(max(0, m^+ - ||V_k||))^2 + \rho(1 - T_k)(max(0, ||V_k|| - m^-))^2 \quad (4)$$

where $k$ represents the $k$-th class, $V_k$ is the output of the DigitCaps and Avepooling layers, and $T_k$ represents the predicted results by the softmax function. That is, $T_k$ equals 1 if the sample is assigned to the $k$-th emotion class, and 0 otherwise. $m^+$, $m^-$ and $\rho$ are the hyperparameters, and they are set to 0.9, 0.1 and 0.5 respectively in this study.

## 2.5 Corpus Adaptation Adversarial Module

Different corpora contain unique characteristics, such as distinct environments or different accents. By discovering the commonness in human emotion, the difference in these unique characteristics can be bridged. The MDD-based CAAM is proposed to identify the cross-corpus emotion by mining the domain-invariant emotion features, so as to discover the feature subspace where the source corpus $P$ and the target corpus $Q$ can be aligned. In CAAM, the model is trained on a labeled sample set containing $n$ samples drawn from the source corpus distribution, $\hat{P} = \{(x_i, y_i)\}_{i=1}^{n}$. An unlabeled sample set with $m$ samples drawn from the target corpus distribution, $\hat{Q} = \{(x_i)\}_{i=1}^{m}$.

As is depicted in Fig.1, CAAM contains our proposed CPAC and two output modules. The CPAC acts as the feature extractor $\psi$ to encode the MFCC features $x^s$, $x^t$ from the source and target corpus, and then the high-level features $\psi(x^s)$, $\psi(x^t)$ are fed to the two output modules. The upper output module $f$ contains one fully-connected (FC) layer of five dimensional neurons to classify speech emotions with the cross-entropy loss function $L(\hat{P}; f)$, which calculates the loss only in the source corpus by:

$$L(\hat{P}; f) = \mathbb{E}_{(x^s, y^s) \sim \hat{P}} L(f(\psi(x^s)), y^s)$$
$$= -\frac{1}{n} \sum_{i=1}^{n} y^s \log(f(\psi(x_i^s))) \quad (5)$$

Here $n$, $f(\psi(x^s))$ and $y^s$ are the number of training examples, outputs of $f$ with the softmax operation and target labels of training samples respectively.

The lower output module $f'$ also contains the same FC layer to reduce the discrepancy in different corpora, and the MDD loss function $D_\gamma(\hat{P}, \hat{Q}; f')$ is employed to calculate the loss between the labeled dataset from source corpus and the unlabeled dataset from target corpus, as defined by:

$$D_\gamma(\hat{P}, \hat{Q}; f') = \text{disp}_{x^t \sim \hat{Q}}(f, f') - \gamma \text{disp}_{x^s \sim \hat{P}}(f, f')$$
$$= \mathbb{E}_{x^t \sim \hat{Q}} L'(f'(\psi(x^t)), f(\psi(x^t)))$$
$$- \gamma \mathbb{E}_{x^s \sim \hat{P}} L(f'(\psi(x^s)), f(\psi(x^s)))$$
$$= \frac{1}{m} \sum_{i=1}^{m} f(\psi(x_i^t)) \log(1 - f'(\psi(x_i^t)))$$
$$- \gamma \left( -\frac{1}{n} \sum_{i=1}^{n} f(\psi(x_i^s)) \log(f'(\psi(x_i^s))) \right) \quad (6)$$

where $\gamma$ is a positive hyperparameter larger than 1, named margin factor. $\text{disp}_{x^t \sim \hat{Q}}(f, f')$ and $\text{disp}_{x^s \sim \hat{P}}(f, f')$ are two margin disparities of the source and target corpus respectively. $f'$ is trained to maximize the distribution discrepancy between two corpora, and $f, \psi$ is trained to minimize the maximum MDD. The objective of adversarial learning for extracting corpus-confused features is formulated as follows:

$$\min_{f, \psi} \max_{f'} (L(\hat{P}; f) + \eta D_\gamma(\hat{P}, \hat{Q}; f')) \quad (7)$$

where $\eta$ is the trade-off coefficient between the cross entropy loss function $L(\hat{P}; f)$ and the MDD loss function $D_\gamma(\hat{P}, \hat{Q}; f')$. And $\eta = 1, \gamma = 1.5$ in this study.

## 3 Experiments
### 3.1 Experiment Setup

**Dataset.** In the experiments, four datasets in different languages are employed, including the Institute of Automation of Chinese Academy of Sciences (CASIA), Berlin Emotional dataset (EmoDB), Surrey Audio-Visual Expressed Emotion dataset (SAVEE), and Ryerson Audio-Visual dataset of Emotional Speech and Song (RAVDESS). The four datasets include 1200, 535, 480, and 1440 data. Their details are given in Tables S1 and S2 in the Supplementary Materials.

**Feature Extraction.** In this experiment, 39-dimensional MFCCs are extracted from the Librosa toolbox [McFee *et al.*, 2015] to serve as the inputs with a frame shift of 0.0125 s and a frame length of 0.05 s.

**Implementation and Training.** The proposed algorithm is implemented with TensorFlow and optimized by using Adam algorithm [Kingma and Ba, 2014]. Especially, the gradient reversal layer (GRL) [Ganin and Lempitsky, 2015] is employed to train $\psi$ to minimize the MDD loss function. Besides, all experimental results in the single-corpus task are based on the 10-fold cross-validation. Details about the training parameters are listed in Section C in the Supplementary Materials.

**Evaluation Metrics.** The weighted average recall (WAR) and the unweighted average recall (UAR) are adopted for performance comparisons. The former refers the mean of recall for different emotional classes and the latter is the weighted mean of recall with weights equal to class probability. Their definitions are given in Equations S1-S2 in the Supplementary Materials.

### 3.2 Comparison to State-of-the-art

**The Single-Corpus Task.** As there are various state-of-the-art methods proposed for different datasets, we select three representative SER methods reporting the top three results on each dataset for the performance comparisons. Consequently, there are different algorithms employed as baselines for different datasets.

As is shown in Table 1, compared to other approaches, CTL-MTNet improves WAR by +4.85%, +3.65%, +0.83%, and +3.40% on CASIA, EMODB, SAVEE, and RAVDESS datasets, respectively. The results confirm that CPAC can establish the sentiment link among different capsules and provide a more robust representation for the SER task.

**The Cross-Corpus Task.** Our method is compared with some representative methods: CDAN [Long *et al.*, 2017], DANN [Abdelwahab and Busso, 2018], and NMFTSL [Luo and Han, 2020]. The results in Table 2 show that our method outperforms all these methods with obvious advantages in all cases. The average UAR and WAR obtained from our method reach 39.90% and 41.57%, respectively. As is shown in Table 2, our method achieves +8.24% and +10.96% improvements of the average UAR and WAR compared to CDAN, +9.50% and +12.28% compared to DANN, and +9.78% and +11.63% compared to NMFTSL. The excellent performance on the cross-corpus datasets verifies that our method can better fit the environments of various accents and cross-languages.

| | Reference | Methods | WAR | | Reference | Methods | WAR |
|---|---|---|---|---|---|---|---|
| CASIA | [Sun et al., 2019] | SVM | 85.08 | EMODB | [Yildirim et al., 2021] | SVM | 87.66 |
| | [Gao et al., 2019] | CNN | 87.90 | | [Tuncer et al., 2021] | SVM | 90.09 |
| | [Hong et al., 2020] | LCNN | 83.65 | | [Ilyas, 2021] | CNN | 91.32 |
| | Our approach | CPAC | **92.75** | | Our approach | CPAC | **94.97** |
| | Reference | Methods | WAR | | Reference | Methods | WAR |
| SAVEE | [Hajarolasvadi and Demirel, 2019] | 3D-CNN | 81.05 | RAVDESS | [Kwon, 2021] | CNN | 85.00 |
| | [Mustaqeem and Kwon, 2021] | CNN | 82.00 | | [Tuncer et al., 2021] | SVM | 87.43 |
| | [Tuncer et al., 2021] | SVM | 84.79 | | [Ibrahim et al., 2021] | ESN | 74.54 |
| | Our approach | CPAC | **85.62** | | Our approach | CPAC | **90.83** |

Table 1: The performance comparisons on CASIA, EMODB, SAVEE and RAVDESS datasets.

| | Source Corpus | | | CASIA | EMODB | CASIA | SAVEE | EMODB | SAVEE | Average |
|---|---|---|---|---|---|---|---|---|---|---|
| | Target Corpus | | | EMODB | CASIA | SAVEE | CASIA | SAVEE | EMODB | |
| [Long et al., 2017] | CDAN | | UAR | 43.17 | 25.60 | 31.67 | 30.40 | 28.17 | 30.95 | 31.66 |
| | | | WAR | 37.25 | 25.60 | 33.61 | 30.40 | 26.67 | 30.15 | 30.61 |
| [Abdelwahab and Busso, 2018] | DANN | | UAR | 46.13 | 25.90 | 26.67 | 29.50 | 25.67 | 28.58 | 30.41 |
| | | | WAR | 39.95 | 25.90 | 25.00 | 29.50 | 29.17 | 26.23 | 29.29 |
| [Luo and Han, 2020] | NMFTSL | | UAR | 30.25 | 28.50 | 34.79 | 33.88 | 24.17 | 29.14 | 30.12 |
| | | | WAR | 29.21 | 28.50 | 34.67 | 33.88 | 24.52 | 28.89 | 29.95 |
| Our approach | CAAM | | UAR | **54.58** | **39.60** | **40.17** | **34.60** | **34.33** | **36.14** | **39.90** |
| | | | WAR | **59.56** | **39.60** | **49.44** | **34.60** | **29.72** | **36.52** | **41.57** |

Table 2: The performance comparisons on the CASIA, EMODB, and SAVEE datasets. The results of RAVDESS are given in Table S3 in the Supplementary Materials

| Method | CNN-Pooling | Self-Attention | CapsNet | CASIA | EMODB | RAVDESS | SAVEE | Average |
|---|---|---|---|---|---|---|---|---|
| Algorithm 1 | No | Yes | Yes | 83.42 | 82.84 | 72.57 | 75.00 | 78.46 |
| Algorithm 2 | Yes | No | Yes | 89.42 | 89.90 | 82.15 | 74.58 | 84.01 |
| Algorithm 3 | Yes | No | No | 84.74 | 82.05 | 76.46 | 69.79 | 78.26 |
| CPAC | Yes | Yes | Yes | **92.75** | **94.97** | **90.83** | **85.62** | **91.04** |

Table 3: The WAR results of the ablation study on the single-corpus task

| Method | MDD | Source Corpus | | CASIA | EMODB | CASIA | SAVEE | EMODB | SAVEE | Average |
|---|---|---|---|---|---|---|---|---|---|---|
| | | Target Corpus | | EMODB | CASIA | SAVEE | CASIA | SAVEE | EMODB | |
| Algorithm 4 | No | | UAR | 39.83 | 24.80 | 25.83 | 26.40 | 25.50 | 25.26 | 27.94 |
| | | | WAR | 33.59 | 24.80 | 23.61 | 26.40 | 25.83 | 25.25 | 26.58 |
| CAAM | Yes | | UAR | **54.58** | **39.60** | **40.17** | **34.60** | **34.33** | **36.14** | **39.90** |
| | | | WAR | **59.56** | **39.60** | **49.44** | **34.60** | **29.72** | **36.52** | **41.57** |

Table 4: The ablation study on the cross-corpus task. The results of RAVDESS are given in Table S4 in the Supplementary Materials

### 3.3 The Ablation Study

To verify the importance of each component of the proposed method, we conduct experiments in the following four aspects:

(1) Algorithm 1 uses the convolutional network provided by the original CapsNet for feature learning;

(2) Algorithm 2 removes the Self-Attention mechanism from the CPAC architecture;

(3) Algorithm 3 replaces the capsule layer with LSTM to extract high-level features;

(4) Algorithm 4 removes the domain adaptation method for the target corpus and trains only with the source corpus.

The first three algorithms are adapted for the single-corpus task, and the last one is used for the cross-corpus task. As is shown in Table 3, on average, our proposed CPAC obtains a relative improvement of +12.58% on WAR compared with the results of Algorithm 1 using only the Conv-Pool component and +7.03% compared with Algorithm 2. These results validate the effectiveness of the CPAC module. Compared to Algorithm 3, applying the CapsNet-based module can beat the LTSM-based module by achieving +12.78% higher score. Similar results can be observed on all datasets, indicating the contribution of different components.

To verify the generalization ability of the CPAC module, we visualize the higher-order features in the test set using Al-gorithms 1-3 based on the t-distributed stochastic neighbor embeddings (t-SNE) technique [Van der Maaten and Hinton, 2008]. As are shown in Fig.2(a) and 2(b), CPAC provides larger and clearer margins among different classes in the mapped feature subspace. Compared with the results of Algorithm 1 in Fig.S2(a) in the Supplementary Materials, CPAC effectively improves the discriminability of the Happy class. Compared with Algorithm 2, CPAC extracts the sentiment features in the audio more selectively, and the clustering effect of each category is tighter. Compared with Algorithm 3 in Fig.S2(c), CPAC corrects the misclassified samples in the Angry, Happy and Surprise classes to a large extent. Therefore, the CPAC structure has excellent generalization ability on a single corpus, which lays the foundation for excellent performance on the cross-corpus dataset.

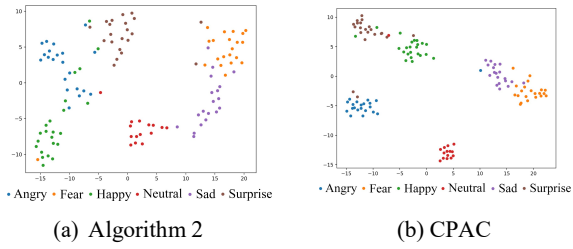

(a) Algorithm 2    (b) CPAC

Figure 2: t-SNE visualization of the high-level features obtained by the Algorithm 2 on CASIA. The results of other ablation algorithms are given in Figure S2 in the Supplementary Materials

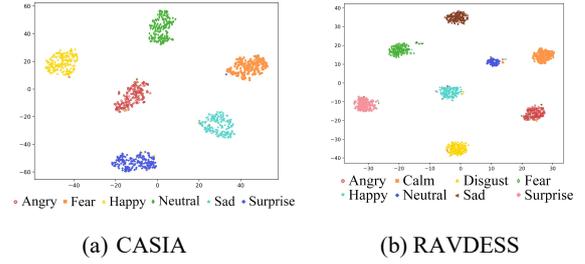

(a) CASIA    (b) RAVDESS

Figure 3: t-SNE visualization of the high-level features on the CASIA and RAVDESS datasets. The results of other datasets are given in Figure S3 in the Supplementary Materials

As is shown in Table 4, compared with Algorithm 4, our algorithm achieves higher performance in all cases with the aid of MDD, gaining +11.96% and +14.99% improvements for the average UAR and WAR indices, and promotes the results to +14.85% and +25.97% higher in UAR and WAR scores in the best cases.

### 3.4 Visualization of Speech Representations

The t-SNE technique is employed to visualize the representations learned by CPAC from different aspects, including distinctive emotional features and domain-invariant emotion features.

**The Single-Corpus Task.** To verify the discriminative ability of our algorithm, Fig.3(a)-3(b) illustrates the 2D visualization projections of the representations extracted from the last fully-connected layer. It is evident that different classes are clustered with clear boundaries in all datasets. The large distances among various classes with few mixed samples guarantee high accurate results, confirming that the proposed CPAC learns the distinctive emotional features well. In contrast, the exception cases shown in Fig.3(a) indicate the proximity between the Fear and Sad emotion classes in CASIA, which leads to the relatively high misclassified cases. As is given in Fig.S1 in the Supplementary Materials, there are 4.5% and 8% samples in the Fear and Sad classes misclassified to each other.

**The Cross-Corpus Task.** The visualized embedding of common sentiment representations in Fig.S4(b) and Fig.S4(c) shows that the proposed CAAM yields superior performance in feature alignment between source corpus and target corpus. As is shown in Fig.S4(a), the distributions of original MFCC features from CASIA and EMODB datasets are not related, in which the average Euclidean distance between them in the same emotion class reaches 2286.44. On the contrary, the high-level features extracted by CAAM show a tighter representation of different corpora in Fig.S4(b) and Fig.S4(c). The average Euclidean distance between them is reduced to 19.87 and 20.53, respectively. Especially in Fig.S4(b), the feature alignment provides higher contributions. It shows that our method can capture domain-invariant sentiment features between different corpora. Moreover, the distance of the same category of emotions is significantly reduced, and the emotions of different categories still maintain certain discrepancies after the feature alignments. However, the separation between classes in the target corpus is not clear enough, which leads to reduced accuracy.

## 4 Conclusion

Speech Emotion Recognition with source and target data from different corpora containing different speakers and languages still remains a big challenge. In this paper, a novel SER algorithm is proposed to tackle the single-corpus and cross-corpus tasks simultaneously. Through the construction of CNN-Pooling and Self-Attention CapsNet, the CPAC model can establish the sentiment connection among different capsules, generating more constant high-level features. Furthermore, the CAAM module by combining CPAC and two MDD algorithms is designed to create more discriminant features and common representation, thus reducing the gap between source and target corpus space. Extensive experiments on four SER datasets validate the stability and superiority of the proposed CTL-MTNet model, with significant improvements over the state-of-the-art algorithms in both tasks.

## Acknowledgments

This work is supported by the National Natural Science Foundation of China (No. 61772023) and Fujian Science and Technology Plan Industry-University-Research Cooperation Project (No.2021H6015). The algorithm production is supported by the AutoDL.com platform.